\documentclass{article}

\usepackage{arxiv}

\usepackage[utf8]{inputenc} 
\usepackage[T1]{fontenc}    
\usepackage{hyperref}       
\usepackage{url}            
\usepackage{booktabs}       
\usepackage{amsfonts}       
\usepackage{nicefrac}       
\usepackage{microtype}      
\usepackage{wrapfig}
\usepackage{lipsum}         
\usepackage{graphicx}
\usepackage{natbib}
\usepackage{doi}
\usepackage{amsmath, bm}
\usepackage{amsfonts}
\usepackage{amssymb}
\usepackage{cleveref}       
\usepackage[english]{babel}
\usepackage{amsthm}
\usepackage{multirow}
\usepackage{tabularx}

\theoremstyle{definition}

\theoremstyle{remark}

\title{Modeling the evolution of temporal knowledge graphs with uncertainty}

\author{ \href{https://orcid.org/0000-0000-0000-0000}{\includegraphics[scale=0.06]{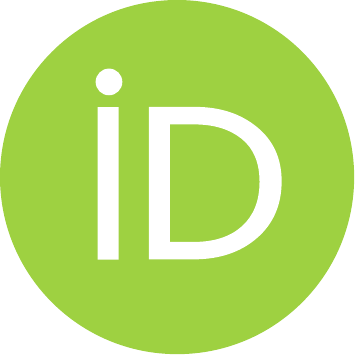}\hspace{1mm}Soeren Nolting} \\
	Department of Computer Science\\
	Ludwig-Maximilian University\\
	Munich, Germany \\
	\texttt{s.nolting@campus.lmu.de} \\
	\And
	\href{https://orcid.org/0000-0000-0000-0000}{\includegraphics[scale=0.06]{orcid.pdf}\hspace{1mm}Zhen Han} \\
	Department of Computer Science\\
	Ludwig-Maximilian University\\
	Munich, Germany \\
	\texttt{zhen.han@campus.lmu.com} \\
	\And
	\href{https://orcid.org/0000-0000-0000-0000}{\includegraphics[scale=0.06]{orcid.pdf}\hspace{1mm}Volker Tresp} \\
	Department of Computer Science\\
	Ludwig-Maximilian University\\
	Munich, Germany \\
	\texttt{volker.tresp@siemens.com} \\
}

\renewcommand{\shorttitle}{Nolting et al.}

\hypersetup{
	pdftitle={Modeling the evolution of temporal knowledge graphs with uncertainty},
	pdfsubject={q-bio.NC, q-bio.QM},
	pdfauthor={Soeren Nolting},
	pdfkeywords={Temporal Knowledge Graphs, Link Prediction, Gaussian Processes},
}

\begin{document}
	\maketitle
	
	\begin{abstract}
		Forecasting future events is a fundamental challenge for temporal knowledge graphs (tKG). As in real life predicting a mean function is most of the time not sufficient, but the question remains how confident can we be about our prediction? Thus, in this work, we will introduce a novel graph neural network architecture (WGP-NN) employing (weighted) Gaussian processes (GP) to jointly model the temporal evolution of the occurrence probability of events and their time-dependent uncertainty. Especially we employ Gaussian processes to model  the uncertainty of future links by their ability to predict predictive variance. This is in contrast to existing works, which are only able to express uncertainties in the learned entity representations. Moreover, WGP-NN can model parameter-free complex temporal and structural dynamics of tKGs in continuous time. We further demonstrate the model's state-of-the-art performance on two real-world benchmark datasets. 
	\end{abstract}

	\keywords{Temporal Knowledge Graphs \and Link Prediction \and Gaussian Processes}

	\section{Introduction}
	Our world becomes increasingly complex with more and more inter-dependencies between social, political, and economical actors. What will happen to  international trade relations and supply chain networks if one country invades another? It is an intriguing challenge to find a data-driven answer to these kinds of questions. One approach to model these complex time-dependent inter-dependencies is to embed them into a temporal knowledge graph. \\\\
	However, in an industrial setting making predictions is not sufficient as potential implications not only have a huge economic impact but, for example, difficult to predict counter-measures might affect the financial performance of a company. To support decision-making we also might want to answer the following question: \emph{How certain can we be about our prediction of these events?} Thus our goal is to jointly model the temporal evolution of the event probabilities and their time-dependent uncertainty.\\
	Events can be regarded as point processes, which are generally intricate to study as they only provide a temporally varying, sparse, and noisy intensity function driving the process. Gaussian processes (GP) offer a well-established framework to model the underlying intensity functions. Recently GPs enjoy increasing attention within academia due to their flexible and non-parametric nature enabling continuous inference across time and moreover model the associated variance and covariance.  \\
	Before we introduce temporal knowledge graphs (tKGs), we will briefly revisit semantic knowledge graphs which store facts in a structured multi-relational knowledge base. Here an event is stored as a triplet $(s, p,o)$ with $s$ being the \emph{subject} and $o$ the \emph{object}. $s$ and $o$ can also be regarded as \emph{nodes} in a directed and labeled graph with the label being the \emph{predicate} $p$. Within research one typically encounters two approaches to statistically model KGs: latent feature models \citep{Ma2018HolisticRF, 10.5555/3104482.3104584}, which attempt to learn embeddings of the entities, relations, and graph feature models (e.g. \cite{10.5555/2878914.2878916}) aiming at exploiting the graph structure of the KG.\\
	Howsoever, in real-life applications, one typically encounters that facts are not static but time-dependent. These temporal events will be encoded in the quadruple $(s,p,o,t)$ extending the triplet with the occurrence time of the fact. Furthermore events that are true over a period of time, e.g. \emph{(Angela Merkel, president of, Germany)},  will be discretized into a sequence of temporal events.\\
	Extending knowledge graphs to temporal knowledge graphs raises the need to statistically model the even more complex dynamics and inter-dependencies between entities in the tKG. Handling these events as an event stream poses a numerical challenge as the number of possible event types is $N_p\cdot N_e^2$, with $N_{p}$ and $N_e$ being the number of predicates and entities respectively. Most of the studies focus on reasoning over tKGs by extending entity and relation embeddings by a time-dependent component in a low-dimensional space but typically miss out on considering the concurrence of events (e.g. \cite{ATiSE})
	and the uncertainty of future events.\\
	In this paper we propose a novel deep learning architecture to capture temporal dependencies as well as to model the uncertainties in the prediction of future events on tKGs.
	Essentially our model consists of three modules. First, a neighborhood aggregation model summarizes the relevant information of the neighborhood of an $(s,p)$-pair. We then encode the timely evolution of the neighborhood in a hidden state $\mathbf{h}_t^{s,p}$ serving as input for dense layers from which we sample a set of $N$ pseudo points $[\tau_i, y_i, w_i]^N_{i=1}$ per object. Each point consists of a weight $w$, a log-scaled time difference $\tau$, and a logit $y$. To keep complexity low, we fit one Gaussian process per object using a 2nd order approximation of uncertainty-adapted cross-entropy loss (see \cite{DBLP:journals/corr/abs-1911-05503}). In other words, we are effectively performing a regression from the logit space to the time domain to model the occurrence probability through the mean function of the Gaussian as well as the associated uncertainty in the prediction by its variance.  \\
	\textbf{To summarize, our main contributions are:}
	\begin{itemize}
		\item We propose a (Weighted) Gaussian process neural network (WGP-NN) predicting future events on real-life large scale tKG, which captures the underlying temporal dynamics of temporal knowledge graphs in \emph{continuous} time. This is in contrast to most of the studies which use discretized state spaces. This enables us to compute the occurrence probability at any point in time.
		\item By using Gaussian processes, WGP-NN naturally can model not only the evolution of the occurrence probability but also the associated \emph{uncertainty} by its variance. 
		\item Moreover the model also satisfies the \emph{locality} property, namely that the uncertainty outside the observed time horizon is higher, especially for $t\to \infty$.
		\item Furthermore, WGP-NN is a non-parametric model enhancing its ability to express complex temporal dynamics.
		\item In contrast to most of the existing models in the literature, WGP-NN natively models the concurrence of events via its Neighborhood aggregation module.
		\item We further demonstrate that WGP-NN achieves more accurate results than state-of-the-art prediction models.
	\end{itemize}
	
	\section{Background and related work}
	\label{sec:background}
	In this section, we will give a brief overview of the underlying theoretical foundations of our work, that is, Gaussian processes as well as temporal knowledge graph prediction. Furthermore, we will introduce related works and the notation used throughout this paper.
	\subsection{Gaussian Process}
	\label{sec:background_gaussian}
	Over the past years, Gaussian Processes have sparked increasing interest within the machine-learning community for various applications (e.g. see (\cite{3569}) or \citealp{https://doi.org/10.48550/arxiv.2009.10862}), especially due to their non-parametric and flexible nature. Formally we can define a Gaussian process like the following:\\
	A Gaussian Process is a collection of random variables satisfying the property that any \emph{finite} subset of those have a joint Gaussian distribution.  Essentially we can think of a Gaussian process as an extension of Gaussian distributions of vectors to infinite dimensions. With a latent, real-valued function $f$, whose values represent the random variables, a Gaussian process can be fully described by its mean $m$ and covariance $k$
	\begin{equation}
		f(\mathbf{x}_i) \sim \mathcal{GP}(\mu(\mathbf{x}_i), k(\mathbf{x}_i, \mathbf{x}_j)),
	\end{equation}
	with $\mu(\mathbf{x}_i) = \mathbb{E}[f(\mathbf{x}_i)]$ and $k(\mathbf{x}_i, \mathbf{x}_j) = [(f(\mathbf{x}_i)-\mu(\mathbf{x}_i))(f(\mathbf{x}_j)-\mu(\mathbf{x}_j))]$. In the regression case, we consider an input data set $\mathbf{X}$ with outputs $\mathbf{f}$ and a test data set $\mathbf{X_*}$ with to-be-predicted outputs $\mathbf{f_*}$ and associated output set $\mathbf{Y}=[\mathbf{y}_i]_{i=1}^N$. Gaussian processes can now be used as prior over the values of $\mathbf{f}=[f(\mathbf{x}_i)]_{i=1}^N$: $	P(\mathbf{f}|\mathbf{X}) = \mathcal{N}(\mathbf{f}|\bm{\mu}, \mathbf{K})$, with $\bm{\mu} = [\mu(\mathbf{x}_i)]_{i=1}^N$ and $K_{ij} = k(\mathbf{x}_i, \mathbf{x}_j)$ respectively. With the compact notation $\mathbf{K} = K(\mathbf{X}, \mathbf{X})$,  $\mathbf{K}_* = K(\mathbf{X}, \mathbf{X}_*)$ and $\mathbf{K}_{**} = K(\mathbf{X}_*, \mathbf{X}_*)$ the joint distribution of $\mathbf{f}$ and the predicted outputs $\mathbf{f_*}$ reads:
	\begin{equation}
		\begin{bmatrix}
			\mathbf{f}\\
			\mathbf{f}_*
		\end{bmatrix} \sim \mathcal{N}
		\bigg(
		\begin{bmatrix}
			m(\mathbf{X})\\
			m(\mathbf{X_*})
		\end{bmatrix},
		\begin{bmatrix}
			\mathbf{K} & \mathbf{K}_*\\
			\mathbf{K}^\top_*¸ & \mathbf{K}_{**}
		\end{bmatrix}
		\bigg),
	\end{equation}
	thus the conditional distribution over $\mathbf{f}_*$ is $\mathbf{f}_*|\mathbf{f},\mathbf{X},\mathbf{X}_* \sim \mathcal{N}\big(\mathbf{K}^\top_*\mathbf{K}\mathbf{f}, \mathbf{K}_{**} - \mathbf{K}^\top_*\mathbf{K}^{-1}\mathbf{K}_* \big)$.\\
	The choice of the kernel function $k(\mathbf{x}, \mathbf{x'})$ is essential in Gaussian process regression as it determines how well the model is able to generalize. In our model, we will resort to the popular squared exponential kernel:
	\begin{equation}
		k(\mathbf{x}_i, \mathbf{x}_j) = e^{(-\gamma^2(\mathbf{x}_i-\mathbf{x}_j)^2)},
	\end{equation}
	where $\gamma$ is a free parameter, characterizing the typical length scale. However, inspired by \cite{DBLP:journals/corr/abs-1911-05503}, we will use an adapted version of the kernel to include weights. As usual, computing the log marginal likelihood gives us the set of optimal hyperparameters $\bm{\Theta}^* = \text{arg}\max_{\bm{\Theta}} \log p(\mathbf{y} | \mathbf{X}, \bm{\Theta})$. 
	\subsection{Future Link Prediction in tKGs}
	A temporal knowledge graph (tKG) $\mathcal{G}$ is a directed, labeled graph with a time attribute associated with any edge. Typically the nodes in a tKG are referred to as \emph{entities} and the edge labels are called \emph{predicates}. Each edge in a tKG can be represented as a quadruple $(e_s, e_p, e_o, t)$. In the literature, the source entity $e_s$ is commonly referred to as subject entity and the target entity $e_o$ is called an object. As both are nodes, they originate from the same set $e_s,e_o \in \{1,\dots, N_e\} = \mathcal{E}$, respectively the predicates $e_p \in \{1,\dots, N_p\}$. \\
	A complementary view on a tKG focuses on the triplets $(e_s, e_p, e_o)$ as \emph{events} occurring at a certain point in time, $t\in \mathbb{R}^+$. Our model will adopt both viewpoints, e.g. exploiting the graph structure of tKGs as well as considering tKGs as a sequence of time-ordered events, to be more precise $\mathcal{G} = \{e_i = (e_{s_i}, e_{p_i}, e_{o_i}, t_i)\}_{i=0}^{N}$ and the observed time-horizon $t_0\leq\dots\leq t_T$.\\
	A typical task on a temporal knowledge graph is \emph{future link prediction}, which is an adaption of the completion task on regular knowledge graphs. Here given a query $(e_s, e_p, ? ,t_i)$ or $(?, e_p, e_o ,t_i)$ the task is to predict the missing object $e_o$ or subject $e_s$ at some future time $t_i>t_T$. Exemplary, in real-life scenarios this task corresponds to predicting the meeting partner in \emph{(Angela Merkel, meets, ?, 08-20-2020)} with the solution \emph{Emmanuel Macron}.\\
	In the following, we will introduce some related models and important works solving the future link prediction graph on temporal knowledge graphs:
	\begin{itemize}
		\item \emph{Know-Evolve} (\cite{DBLP:journals/corr/TrivediDWS17}): Here the occurrence probability of an event in the tKG is modeled as point process and learns entity embeddings which scale dynamically over time using score functions.
		\item \emph{RE-NET} (\cite{DBLP:conf/emnlp/JinQJR20}): In this approach the famous R-GCN model. (\cite{https://doi.org/10.48550/arxiv.1703.06103}) is used to summarize entity neighborhoods and uses the historical sequence of past events to forecast the future. 
		\item \emph{ATiSE} (\cite{ATiSE}): Here the authors used additive time series decomposition to scale entity representations. To be more precise the encoding of the embeddings can be split here into a trend, seasonal and a random (noise) component, which parameters are learned using the KL divergence.
		\item \emph{GHNN} (\cite{DBLP:journals/corr/abs-2003-13432}): In the GHNN model the neural hawkes process (\cite{DBLP:journals/corr/MeiE16}) is augmented to tKGs.
		\item \emph{NODE} (\cite{han-etal-2021-learning-neural}): NODE extends the idea of neural ordinary differential equations to capture the structural as well as the temporal information into dynamically evolving entity embeddings.
	\end{itemize}
	Marrying uncertainty and temporal knowledge graphs is a relatively new field of study. \cite{Marrying_uncertainty} extends the idea of Markov Logic Networks to temporal knowledge graphs and incorporates confidence scores to facts to express the associated uncertainty.\\
	However, the task of future link prediction in tKG can also be seen as a multi-class classification task. Here the forecast of asynchronous event sequences has recently enjoyed a gain in popularity, e.g. \cite{DBLP:journals/corr/abs-1911-05503} modeled the (class) probability simplex employing a Dirichlet or logistic normal distribution, which effectively enables them to model also the associated uncertainty of the prediction. 
	\subsection{Notation}
	\label{sec:notation}
	Following the notation of \cite{DBLP:journals/corr/abs-2003-13432}, we will denote an event $e_i$ consisting of the subject, predicate and object tuple at timestamp $t_i$  $(e_{s_i} , e_{p_i}, e_{o_i}, t_i)$. Scalars are denoted in lowercase letters, e.g. $\tau$ the log-normalized time difference. The set of all $e_{s_i}$ and $e_{o_i}$ is denoted by $\mathcal{E}$ and its cardinality $N_e = |\mathcal{E}|$. Vectors are denoted in lowercase and bold, e.g. the embedding vectors of the subject, predicate, object tuple will be written as $\mathbf{e}_{s_i}$, $\mathbf{e}_{p_i}$, $\mathbf{e}_{o_i}$ respectively. Matrices are denoted in bold and uppercase such as the kernel matrix $\mathbf{K}$ of the Gaussian process.
	\begin{figure}
		\centering
		\includegraphics[width=\linewidth]{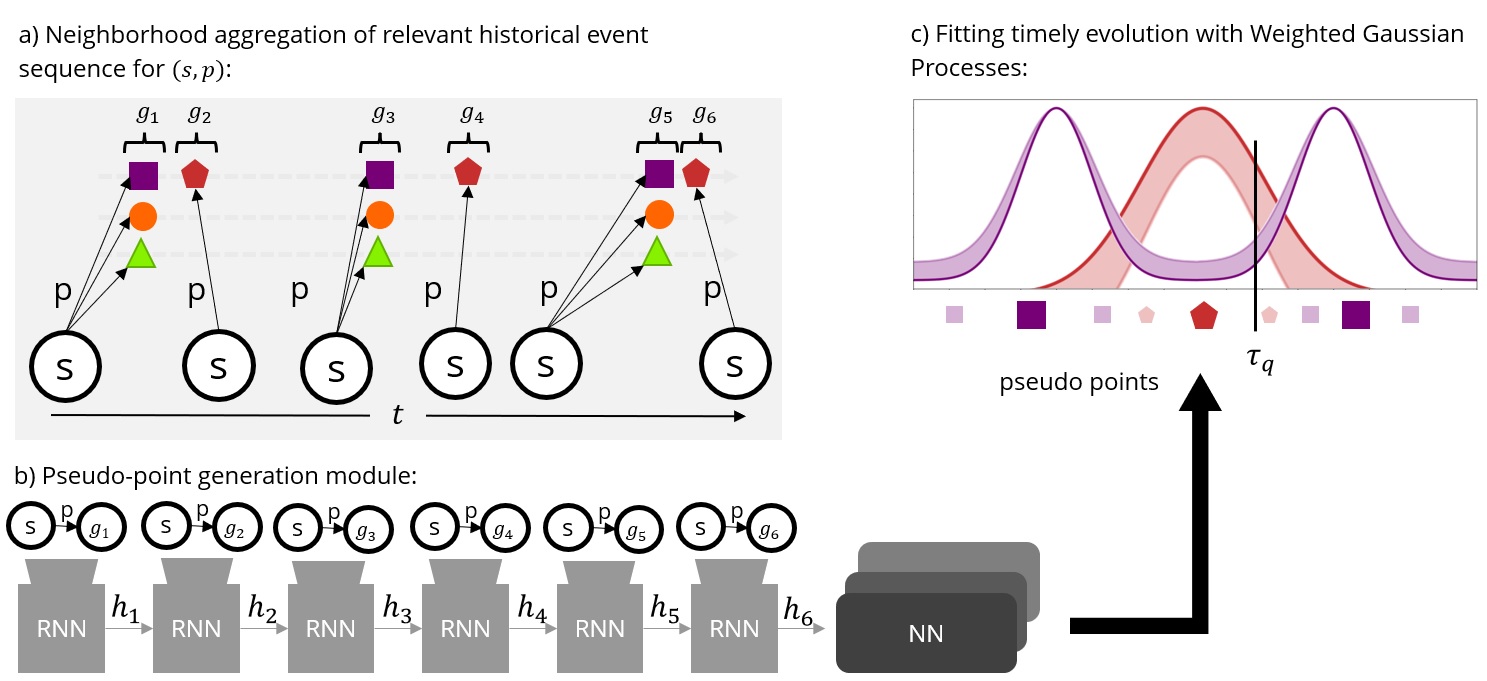}
		\caption{Visual interpretation of WGP-NN's architecture. a) Depicts the Neighborhood aggregation module for a given $(s,p)$-pair. The colored shapes visualize the connected nodes at certain points in time. Past graph slices considered $M = 6$. b) Shows the pseudo-point generation module which encodes the interaction history in the hidden state $\mathbf{h}_t$ serving as input to dense layers generating the pseudo points. The shade of the pseudo points symbolizes their associated weights $w$. c) Visualizes the weighted gaussian process essentially performing a regression from the logit space to the time domain to model the occurrence probability and variance. $\tau_q$ represents an arbitrary (log-normalized) query time.}
		\label{fig:model intuition}
	\end{figure}
	
	\section{Our Method}
	\label{sec:model}
	In this section, we present the Weighted Gaussian Process Neural Network (WGP-NN) for modeling sequences of discrete large-scale multi-relational graphs in continuous time. Especially, we will model the evolution of the occurrence probability of an entity over time with uncertainty. This architecture consists mainly of the following three modules:
	
	\begin{itemize}
		\item \textbf{Neighborhood aggregation module:} This module captures information about neighboring nodes that are connected at a certain point in time. 
		\item \textbf{Pseudo-Point generation module:} A recurrent neural network learns the evolution history of neighboring nodes serving as input for dense layers which generate a set of weighted points.
		\item \textbf{Weighted Gaussian Process:} A weighted Gaussian process is used to model the evolution of the probability of the nodes and their uncertainty.
	\end{itemize}
	A visual interpretation of the model can be found in figure \ref{fig:model intuition}.In the remaining part of this section, we will introduce the modules of WGP-NN in greater detail.  
	\subsection{Relevant Historical Event Sequence}
	We formalize a temporal knowledge graph as a sequence of knowledge graphs $\mathcal{G}=(\mathcal{G}_1, \mathcal{G}_2,\dots,\mathcal{G}_T)$, whereas each slice $\mathcal{G}_{t_i} = ((e_{s} , e_{p}, e_{o}, t)\in \mathcal{G}| t = t_i)$
	is a multi-relational directed graph. Inspired by \cite{DBLP:conf/emnlp/JinQJR20}, we consider events occurring at the same timestamp to belong to the same graph slice. Furthermore, we assume that events observed at $t_{i}$ occur independently from each other given the past observed history of $\mathcal{G}_{n<i}$. We will further imply that the Markov assumption holds i.e., that for the task of predicting an object $(e_{s_i} , e_{p_i}, ? , t_i)$ in the temporal knowledge graph, only $M$ past graph slices need to be taken into account. Thus, the conditional probability to be modeled reads $\mathbb{P}(e_{o_i} | e_{s_i} , e_{p_i}, t_i, \mathcal{G}_{i-1},\dots,\mathcal{G}_{i-M})$.\\
	Inspired by \cite{DBLP:journals/corr/abs-2003-13432}, we reduce the complexity of our model by the assumption that not all events at past graph slices need to be taken into account to evaluate the aforementioned conditional probability. Instead, only a subset of events $\mathbf{O}$, namely those which formed a link to $(e_{s_i}, e_{p_i})$ at past slices needs to be considered. More formally we define the past event sequence $e_i^{h,s,p} = \{\bigcup_{t_{i-M}\leq t_j < t_i}(e_{s_i}, e_{p_i}, \mathbf{O}_{t_j}(e_{s_i}, e_{p_i}))\}$. Thus the conditional probability of an object $e_o$ occurring at time $t_i$ reads now 
	\begin{equation}
		\mathbb{P}(e_{o_i} | e_{s_i} , e_{p_i}, t_i, \mathcal{G}_{i-1},\dots,\mathcal{G}_{i-M})= \mathbb{P}(e_{o_i} | e_{s_i} , e_{p_i}, t_i, e_i^{h,s,p}).
	\end{equation}
	\subsection{Neighborhood aggregation}
	Within a graph slice $\mathcal{G}_{t_j}$ a subject $e_{s_i}$ with a predicate $e_{p_i}$ can form multiple links. Thus, we employ the well-known element-wise mean aggregation (\cite{DBLP:journals/corr/HamiltonYL17}) over all embedding vectors at each point in time over the subset of connected objects $\mathbf{O}_{t_j}(e_{s_i}, e_{p_i})$:
	\begin{equation}
		g\left(\mathbf{O}_{t_j}(e_{s_i}, e_{p_i})\right) = \left(|\mathbf{O}_{t_j}(e_{s_i}, e_{p_i})|\right)^{-1}\sum_{e_o\in\mathbf{O}_{t_j}(e_{s_i}, e_{p_i})}\mathbf{e_o}
	\end{equation}
	
	\subsection{Pseudo-point generation}
	Pseudo-points in Gaussian process regression are typically used to reduce computational complexity e.g. see \cite{NIPS2005_4491777b}. This will be especially important for controlling training costs on real-world tKGs. On the other side, it also allows our model to assign weights to each generated point and by this judge the importance of each point w.r.t. to the considered subject, predicate tuple, and their history in the event sequence. To summarize, each point consists of a weight $w$, a log-scaled time difference $\tau$ and a logit $y$. In the end, the set of generated points will characterize the probability distribution 
	$\mathbb{P}(e_{o_i} | e_{s_i} , e_{p_i}, t_i, e_i^{h,s,p})$\\
	To generate a set of the $N$ pseudo points $\mathcal{X}^{(e)} = \big[\tau^{(e)}_j, y^{(e)}_j, w^{(e)}_j\big]^N_{j=1}$ per entity $e \in \mathcal{E}$ in the tKG $\mathcal{G}$, we first learn the complex time-dependencies via encoding the interaction history of the tuple over time via an RNN (\cite{DBLP:journals/corr/ChoMGBSB14}). Then we draw the pseudo-points from its hidden state $\mathbf{h}$ using dense layers:
	\begin{eqnarray}
		\mathbf{h}_t &&= \text{RNN}(\mathbf{e}_{s_i},\mathbf{e}_{p_i}, g(\mathbf{O}_{t_j}(e_{s_i}, e_{p_i})), \mathbf{h}_{t-1}),\\
		\bm{\tau} &&= f_\text{splus}(\text{NN}_\tau(\mathbf{h}_t)),\\
		\mathbf{y} &&=  f_\text{ReLU}(\text{NN}_y(\mathbf{h}_t)),\\
		\mathbf{w} &&= f_\text{sigm}(\text{NN}_w(\mathbf{h}_t)),
	\end{eqnarray} 
	where $f_\text{splus}$ denotes the element-wise softplus-activation function $f_\text{splus}(x) = \log(1+\exp(x))$, the element-wise rectifier linear unit activation function $f_\text{ReLU}(x)=\max(0,x)$ and the element-wise sigmoid function $f_\text{sigm} = (1 + \exp(-x))^{-1}$.\\
	In other words, the pseudo-point generation can be seen as an encoder-decoder architecture in which the model first encodes the relevant interaction history of a subject/predicate-pair but then decodes to a pseudo tKG which will be optimized for predicting the evolution of the real tKG. 
	\subsection{Weighted Gaussian Process}
	Given a relevant historical event sequence for a subject and predicate, we aim to model the evolution of the occurrence probability and the associated uncertainty of the next object over time. We treat the next objects as a random variable, which we consider as values of a (generating) function $f$. WGP-NN models this function now as a Gaussian process
	\begin{equation}
		f(\mathbf{x}) \sim \mathcal{GP}(\mathcal{X}, \bm{\Theta}),
	\end{equation}
	where the parameter $\bm{\Theta}$ incorporates the mean function $\bm{\mu}_\epsilon(\tau)$ and  the variance $\mathbf{\Sigma}_\epsilon$. The values of $f$ are the object logits $\bm{y}$ and the inputs $\bm{x}$ are the log-scaled time-differences $\bm{\tau}$ and the weights $\mathbf{w}$ of our pseudo-point generation module. Our conditional probability changes accordingly to 
	\begin{equation}
		\label{eq:cond_prob}
		\mathbb{P}(e_{o_i} | e_{s_i} , e_{p_i}, t_i, e_i^{h,s,p}) \equiv  \mathbb{P}(e_{o_i} | \mathcal{X}, \bm{\Theta})
	\end{equation}
	With the background from section \ref{sec:background_gaussian} in mind, the mean and the variance of the Gaussian process $\mathcal{GP}$ can now be modeled as 
	\begin{eqnarray}
		\mu_\epsilon(\tau) &&= \mathbf{k}_\epsilon^T\mathbf{K}^{-1}_\epsilon\mathbf{y}_\epsilon,\\
		\sigma^2_\epsilon(\tau) &&= k(\tau, \tau) - \mathbf{k}_\epsilon^T\mathbf{K}^{-1}_\epsilon\mathbf{k}_\epsilon.
	\end{eqnarray}
	as stated above, inspired by \cite{DBLP:journals/corr/abs-1911-05503} we will use an adapted version of the squared exponential kernel to enable our model to judge the importance of the generated pseudo-points: $k(\tau,\tau') = \min(w, w')\exp(-\gamma^2(\tau-\tau')^2)$.
	\subsection{Parameter learning}
	In the case of future \emph{link prediction}, the task is to predict a future link in the tKG. To be more precise given a object prediction query $(e_{s_i} , e_{p_i}, ?, t_i)$ and the pseudo-points $\mathcal{X}$ and parameters $\Theta$, we would like to find the object with the maximal conditional probability given in Eq. \ref{eq:cond_prob}.\\
	In essence, this problem can be seen as a multi-class classification task, whereas the classes represent the various object candidates. However, the classical cross- entropy loss is an ill-suited measure in our case as our model not only aims to predict the missing object in the query but also its uncertainty. With the true categorical distribution $p_{ic}^*$, the cross-entropy loss is given by $\mathcal{L}^{sp}_{i} = - \sum_{c=1}^{N_e}p_{ic}^*\log(p(e_{o_i}=c|\mathcal{X}, \Theta))$. Due to point estimate $p(e_{o_i}=c|\mathcal{X}, \Theta)$, the uncertainty in the distribution is neglected.
	To circumvent this shortcoming, the loss introduced in \cite{DBLP:journals/corr/abs-1911-05503} is used:
	\begin{align}
		\mathcal{L}_{\text{UCE},i}^{sp} &= - \int P_i(\theta(\tau^*))\sum_{c=1}^{N_e}p_{ic}^*\log(p(e_{o_i}=c|\mathcal{X}, \Theta))+r_c\\
		&\approx \mu_{c_i}(\tau_i^*) - \log\big(\sum_{c=1}^{N_e}\exp(\mu_{c_i}(\tau_i^*)+0.5\sigma_c^2(\tau_i^*))\big) + \frac{\sum_{c=1}^{N_e}\big(\exp(\sigma_c^2(\tau_i^*)-1)\exp(2\mu_{c_i}(\tau_i^*)+\sigma_c^2(\tau_i^*))\big)}
		{2\big(\sum_{c=1}^{N_e}\exp(\mu_{c_i}(\tau_i^*)+0.5\sigma_c^2(\tau_i^*))\big)^2} +r_c.
	\end{align}
	The second expression is a 2nd order series expansion, which is fully back-propagate-able. $r_c$ is a regularization term that ensures that we do not generate points well-outside our observed time scale with high confidence, $r_c = \int_{0}^{T}\alpha(\mu_{c}(\tau))^2+\beta(\nu-\sigma_c^2(\tau))^2d\tau$. The first term pushes the mean to zero, the second term pushes the variance to $\nu$.
	\section{Experiments}
	\label{sec:experiment}
	In this section we report on the conducted experiments on two common tKG  datasets. The performance of WGP-NN is compared against nine benchmarks from literature.
	\subsection{Experimental Setup}
	\subsubsection*{Datasets}
	To evaluate the performance of our model, we use two datasets commonly used in literature for link prediction on tKGs (e.g. see \cite{DBLP:conf/emnlp/JinQJR20}): the YAGO dataset (\cite{DBLP:conf/cidr/MahdisoltaniBS15}) and the Integrated
	Crisis Early Warning System (ICEWS, \citealt{DVN/28075_2015}). Each dataset can be seen as representative of a class of tKG datasets:
	\begin{itemize}
		\item \emph{Event-based} datasets (e.g. ICEWS): Here each quadruple represents an event occurring at a single point in time,
		\item \emph{Period-based} datasets (e.g. YAGO): A triple in a knowledge graph is equipped with a time period $[t_s,t_e]$ within that fact is valid, $(s,p,o,[t_s,t_e])$. Typically these events are then converted into event-based datasets by discretization into a sequence of events.
	\end{itemize} 
	Of course, the transition between these classes is fluid. For our evaluation, we have limited us to the ICEWS18 dataset, which restricts the set to events occurring in 2018.
	\subsubsection*{Evaluation settings and metrics}
	\begin{wraptable}{r}{5.5cm}
		\caption{Grid search: Explored hyperparameter space}
		\label{tab:hyper}
		\begin{tabular}{cc}\\\toprule  
			\hline
			Parameter & Space  \\
			\hline
			Past graph slices $M$ & $\{4,6,8,10\}$ \\  		
			Pseudo points $N$ & $\{1,2,4,6\}$  \\
			Embedding size & $\{200, 300\}$ \\
			Batch Size & $\{600, 800, 1000\}$ \\  
			\hline
		\end{tabular}
	\end{wraptable} 
	For better comparison, we used the time-stamp-wise train(80\%)/valid(10\%)/test(10\%) splitting taken from \cite{DBLP:conf/emnlp/JinQJR20}.¸To be more precise, the splitting performed ensures that $t_{\text{train}}<t_{\text{valid}}<t_{\text{test}}$ for all timestamp-triplets from the train, validation and test set.\\
	Following the definition of the evaluation metrics of \cite{DBLP:journals/corr/abs-2003-13432}, we report the Mean Reciprocal Rank (MRR) as well as HITS@3/10. To find the optimal set of hyperparameters, we have conducted a grid search for each dataset over the parameter space shown in table \ref{tab:hyper}. We chose the set of hyperparameters which showed the best result after 5 iterations. The learning rate of the Adam optimizer (\cite{https://doi.org/10.48550/arxiv.1412.6980}) was set to $lr=10^{-3}$, the regularization parameters $\alpha$ and $\beta$ were fixed to $10^{-3}$. For economical reasons, the experiments were performed on Intel(R) Xeon(R) Platinum 8375C CPUs with RAM memory of up to 384GBs.
	\subsection{Performance Comparison}
	\subsubsection*{Baselines}
	We compare our model against nine baselines, including three static ones: Distmult (\cite{https://doi.org/10.48550/arxiv.1412.6575}), TuckER (\cite{Balazevic_2019}), and COMPGCN (\cite{COMPGCN}). Furthermore, we report\footnote{based on work conducted in \cite{Tango} } on six tKG baselines: TTransE (\cite{10.1145/3184558.3191639}), TA-DistMult (\cite{garcia-duran-etal-2018-learning}), CyGNet (\cite{CyGNet}), DE-SimplE (\cite{Goel_Kazemi_Brubaker_Poupart_2020}), TNTComplEx (\cite{TNT}), RE-NET (\cite{DBLP:conf/emnlp/JinQJR20}). 
	For both data sets, we provide time-aware filtered results, following the evaluation setting of \cite{DBLP:journals/corr/abs-2003-13432}. Here we filter out, at test time, the corrupted triplets occurring at the query time $t_q$ that appear within training, validation, or test set except the one of interest. This is common practice to provide a fair evaluation of the model's performance as otherwise concurrent true prediction at $t_q$ would be punished. The results without filtering, commonly referred as \emph{raw} evaluation setting can be found in appendix \ref{sec:rawresults}.
	\begin{table}
		\centering
		\caption{Future link prediction results for nine benchmarks on two datasets. The best results are marked in bold.}
		\label{tab:results}
		\begin{tabularx}{\textwidth}{|l||XXX|XXX|}
			\hline
			Data set &
			\multicolumn{3}{|c|}{ICEWS18 - time aware} &
			\multicolumn{3}{|c|}{YAGO - time aware}
			\\
			\hline
			Model& MRR & H@3 & H@10& MRR & H@3 & H@10\\
			\hline
			Distmult   & 16.69&18.12&31.21 &54.84&59.81&68.52\\
			TuckER&    20.68 & 22.60 & 37.27& 54.86&59.63&68.96\\
			COMPGCN  & 20.56 &22.96&38.15 &54.35&59.26&68.29\\
			\hline
			TTransE      &8.08 & 8.25 & 21.29 &   31.19&40.91&51.21\\
			TA-DistMult & 11.38 &12.04&22.82 &54.92&59.61&66.71\\
			CyGNet&24.93 &28.28 & 42.61 & 52.07&56.12&63.77 \\
			DE-SimplE &19.30 &21.86& 34.80&54.91 &57.30&60.17\\
			TNTComplEx& 21.23 &24.02 & 36.91 &57.98 &61.33&66.69\\ 
			RE-Net  & 27.90 & 31.37 & \textbf{46.37} &58.02 &61.08&66.29\\
			\hline
			WGP-NN& \textbf{33.16}   & \textbf{35.21}& 45.27 &\textbf{64.97}&\textbf{67.25}&\textbf{74.66}\\
			\hline
		\end{tabularx}
	\end{table}
	\subsubsection*{Results}
		Table \ref{tab:results} shows the evaluation of the (time aware filtered) future link prediction on the ICEWS18 as well as the YAGO dataset. Results demonstrate that WGP-NN outperforms all static baselines, highlighting the importance to consider temporal information in the prediction. Especially, this can be seen for the ICEWS18 results as in this data set shows pronounced temporal dynamics.\\
		Furthermore, WGP-NN outperforms all baselines that are taking time into account, showing the superiority of the chosen approach and the strength of our method to incorporate temporal information in the extrapolated link prediction.
	\section{Conclusion}
	\label{sec:Conclusion}
		We propose a novel graph neural network architecture to forecast future links in temporal knowledge graphs. In contrast to existing models, WGP-NN natively models the structural dependencies e.g. the concurrence of events. Furthermore, our model can predict in a continuous time the occurrence probability of links in the future with their associated variance over time. We test our model on two temporal knowledge graph datasets and demonstrate that it outperforms state-of-the-art results.
	\bibliographystyle{unsrtnat}
	\bibliography{template}  
	\appendix
	\newpage	

	\section{Raw results}
	\label{sec:rawresults}
		\begin{table}
		\centering
		\caption{Raw future link prediction results for six benchmarks on two datasets. No time-aware filtering is applied. The best results are marked in bold.}
		\label{tab:resultsraw}
		\begin{tabularx}{\textwidth}{|l||XXX|XXX|}
			\hline
			Data set &
			\multicolumn{3}{|c|}{ICEWS18 - raw} &
			\multicolumn{3}{|c|}{YAGO - raw}
			\\
			\hline
			Model& MRR & H@3 & H@10&MRR & H@3 & H@10\\
			\hline
			Distmult   & 16.30& 17.67& 30.93 &\textbf{47.66} &55.89&67.45 \\
			TuckER&   20.20&21.99&36.91&47.48&55.55&68.07\\
			COMPGCN &19.98 & 22.25 & 37.73 & 47.08& \textbf{66.90}&\textbf{68.81}\\
			\hline
			TTransE    &7.92 & 8.00 & 21.02  & 26.18&36.16&48.00\\
			TA-DistMult&   11.05 & 11.72 & 22.55  & 45.54&51.08&62.15\\
			RE-Net&  26.62 & 30.26 & \textbf{45.82} &46.28&51.77&61.55\\
			\hline
			WGP-NN& \textbf{31.92}  & \textbf{33.74 }& 43.43 & 47.44&54.47&65.23\\
			\hline
		\end{tabularx}
	\end{table}
Table \ref{tab:resultsraw} shows the future link prediction results with no filtering applied, hence \emph{raw} results. Commonly, one regards this evaluation setting as an unfair setting for tKGs as the models are punished for correct predictions for concurrent links at test time. Nevertheless, WGP-NN outperforms static and tKG baselines for the MRR and Hits@3 metric on the ICEWS18 benchmark due to the pronounced temporal dynamics in this dataset.\\
As in the period-based YAGO dataset the temporal dynamics are comparatively weak, static baselines typically outperform tKG baselines. However, note that WGP-NN still outperforms all tKG baselines. 
\end{document}